Tech Science Press

# Edge Detection and Deep Learning Based SETI Signal Classification Method

## Zhewei Chen[1], Sami Ahmed Haider[*]

[1]Zhejiang Normal University, Jinhua, 321004, China
[*]Zhejiang Normal University, Jinhua, 321004, China
*Corresponding Author: Sami Ahmed Haider. Email: samiahmed@zju.edu.cn

**Abstract**：Scientists at the Berkeley SETI Research Center are Searching for Extraterrestrial Intelligence (SETI) by a new signal detection method that converts radio signals into spectrograms through Fourier transforms and classifies signals represented by two-dimensional time-frequency spectrums, which successfully converts a signal classification problem into an image classification task. In view of the negative impact of background noises on the accuracy of spectrograms classification, a new method is introduced in this paper. After Gaussian convolution smoothing the signals, edge detection functions are applied to detect the edge of the signals and enhance the outline of the signals, then the processed spectrograms are used to train the deep neural network to compare the classification accuracy of various image classification networks. The results show that the proposed method can effectively improve the classification accuracy of SETI spectrums.

**Keywords:** Deep Learning，Edge Detection，Image Processing，SETI

## 1 Introduction

In January 2016, the Berkeley SETI Research Center at the University of Berkeley launched a project called Breakthrough Listening[1], which is described as the most comprehensive search for alien communications to date. Scientists use the Green Bank Observatory's radio telescopes in West Virginia to monitor and collect radio signals from multiple directions in space for signs of extraterrestrial intelligence. During the observation, the amplitude, frequency, phase and other attributes of these signals are recorded by the receiver, then processed and converted into complex-valued time-series data. Since the narrowest bandwidth of the signal emitted by the celestial body itself in the universe is 500 Hz, and the bandwidth of artificial signals is usually much narrower than that emitted by natural processes, SETI focuses on searching for unnatural artificial signals with bandwidth less than 500 Hz. The previous approach used to classify various narrowband signals is by traditional digital signal processing[2]. However, this approach can only detect one kind of narrowband signal whose frequency drifts linearly with time. This approach cannot classify different kinds of narrowband signals and has the problem of low classification accuracy caused by background noise[3]. SETI has therefore developed a machine vision algorithm that converts narrowband signals into two-dimensional time-frequency spectrograms through Fourier Transform. Depending on the contour of radio signal shown in the spectrums, human beings can easily distinguish different signal categories.

With the progress of deep learning algorithms, the improvement of computing power and the development of big data technology in the last two decades, the application of image classification in astronomy has achieved amazing success. In the summer of 2017, SETI held a machine learning challenge to prove the accuracy of this machine vision algorithm, providing participants with signal simulation datasets and test sets of various sizes. The work in this paper will use the small version data set of the competition, which consists of seven different categories of SETI signals, and each category contains a total of 1000 signal data[4]. Because of the different characteristics represented by the different kinds of signals, a single kind of filter is not suitable for denoising all signals at the signal processing level[5]. Therefore, this paper adopts a new method. After the signals are converted into two-dimensional gray spectrums, Gaussian filter and edge detection operator are applied to preprocess them. Then an image classification network will be used to realize the accurate multi classification of SETI signals. In this study, convolutional neural networks and image processing techniques are applied to classify SETI signals, which will be helpful for further research on data mining of astronomy.

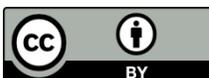

## 2 Related Work

SETI intends to use advanced equipment such as radio telescopes to receive electromagnetic waves from the universe, including background radiation, radio waves from stars and other noise in the galaxy, to analyze regular radio signals in the hope of discovering alien civilizations. G.A.[6] proposed that the observed characteristics of SETI radio signals are usually most obvious in the form of spectrograms. By simulating complex-valued time-series data, a data bank of 140,000 signals from seven different signal categories is generated. Then, by training the wide residual network, the gray two-dimensional spectrograms of these signals are used for training and classification. By Harp G.R.[5], the work of G.A.[6] has been improved in image processing by using vertical and horizontal feature filters and Huff transformations to extract spectral characteristics and enhance the outline of the signal. It is also pointed out that the noise in the background of the spectrograms cannot be reduced by high-pass, low-pass and band-pass filters, and the spectrograms of different kinds of signals have different characteristics, so a single kind of filter cannot be applied to all kinds of signal for noise reduction. In the work of Malgorzata G[7], three different feature datasets are constructed by different preprocessing methods. Combining three different models including DenseNet, the combined network can learn the different hierarchical features of the spectrograms and obtain the classification effect of SETI images in a similar form of integrated learning. However, the research shows that the combined network takes up a large amount of computer memory. Although it is relatively beneficial to improve the classification accuracy, the disadvantages are obvious: loose structure, too many hyperparameters and long training time.

In addition, many international facilities are working with the SETI project. As the largest single-aperture radio telescope in the world, the Five-hundred-meter Aperture Spherical radio Telescope (FAST) has a superior position and is capable of sensitive searching for signals from Extraterrestrial Intelligence. The work of Di[8] revealed some unique characteristics of FAST, and extended SETI observation to three different signal types, namely, narrow band, wide-band artificially dispersed and modulated signals. At the same time, the idea of fast observation has been proposed, and unprecedented observation sensitivity of FAST will help human to discover more advanced alien civilization in the current technological scope. And in the work of Zhang[9], the strategies of SETI synchronous observation were analyzed, then the overall framework of SETI back-end system was validated by processing the FAST drift scan data. The methods of removing radio interference from data and extracting candidate targets were also analyzed.

### 2.1 Simulated Signals

Based on the expertise, SETI simulated six narrowband signals and one kind of background noise to represent the most common set of observed signals[5], which have a narrow bandwidth and the frequency of signals usually varies over time to simulate frequency drift (Doppler drift) caused by the earth's rotation or the movement of the signal source. Narrowband signals with constant frequencies in real observations are likely to come from terrestrial radio stations or geosynchronous satellites and are therefore not concerned. In addition, simulated signals in the dataset are saved in complex numbers form[10] and represented as 8 bits real values and 8 bits image values.

## *2.2 Spectrograms*

The data represented by these complex-valued time series can be transformed from time domain to frequency domain by Fourier transform to generate two-dimensional time-frequency spectrograms. This is because a periodic signal can be decomposed into multiple sine or cosine signals as follows:

$$f(t) = \frac{a_0}{2} + \sum_{n=1}^{\infty}[a_n\cos(n\omega t) + b_n\sin(n\omega t)] \tag{2-1}$$

Because the original periodic function is approximated by the superposition of multiple sine and cosine waves, $f(t)$ represents the approximate value of the signal in the time domain, $\omega$ represents the frequency of the signal. $a_n$ and $b_n$ represent the amplitude of the signal, which can be expressed as:

$$a_n = \frac{2}{T}\int_{t_0}^{t_0+T}f(t)\cos(n\omega t)dt \tag{2-2}$$

$$b_n = \frac{2}{T}\int_{t_0}^{t_0+T}f(t)\sin(n\omega t)dt \tag{2-3}$$

Where, T in the formula represents the signal period. To use Fourier series in the complex space, it needs to use Euler formula:

$$e^{i\theta} = \cos(\theta) + i\sin(\theta) \tag{2-4}$$

$$\cos(\theta) = \frac{e^{i\theta} + e^{-i\theta}}{2} \tag{2-5}$$

$$\sin(\theta) = -i \cdot \frac{e^{i\theta} - e^{-i\theta}}{2} \tag{2-6}$$

Euler formula realizes the conversion between sine, cosine and exponential function, and Fourier series can be represented in complex space as:

$$f(t) = \frac{1}{T}\sum_{n=-\infty}^{+\infty}\int_{t_0}^{t_0+T}f(t)e^{-in\omega t}dt \cdot e^{in\omega t} \tag{2-7}$$

After Fourier transform, the signal spectrums can be generated. The vertical axis shows time, the horizontal axis shows frequency, and the color of the pixel corresponds to the signal power at the corresponding time and frequency. These spectrums are basic images in this classification task.

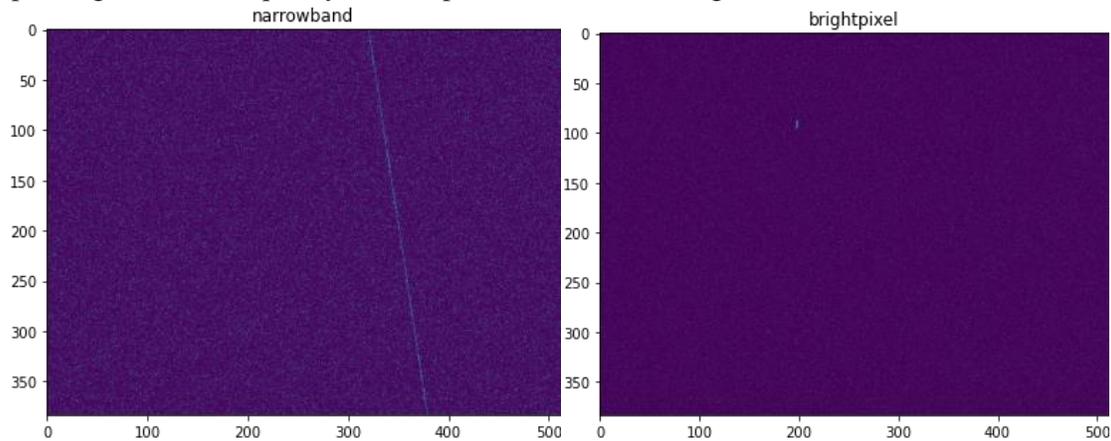

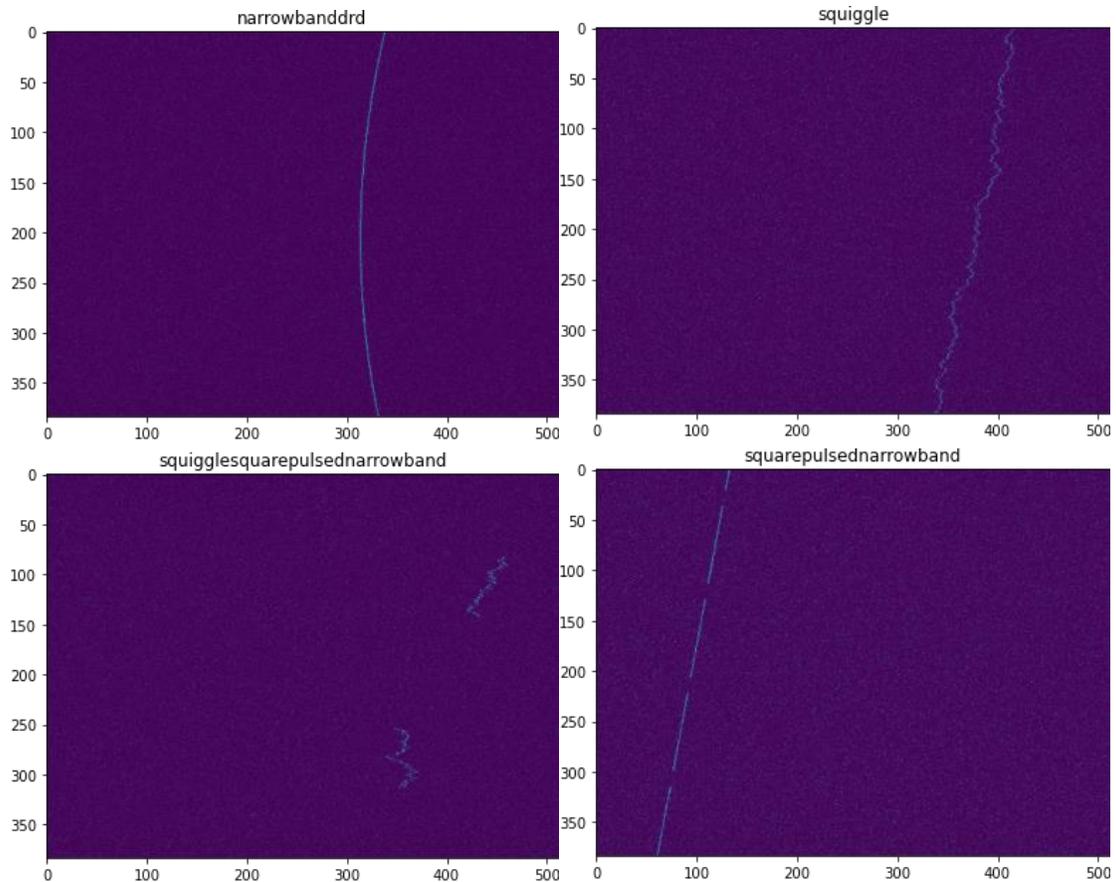

**Figure 1:** Spectrograms of all categories of narrowband signals

The picture above gives an example of spectrums for six kinds of narrowband simulated signals: narrowband, narrowband with curvature (narrowbanddrd), narrowband with square wave amplitude modulation (squarepulsenarrowband), brightpixel, squiggle, squiggle with square wave amplitude modulation (squigglesquarepulsednarrowband). In addition, there is another kind of signal labeled as noise doesn't list above, whose background only has Gaussian white noise. The simulation process and parameters of each kind of signals have been explained in detail by[6].

## *2.3 Edge Detection*

The significant changes in the images usually reflect the important events and changes of attributes. The purpose of edge detection is to identify the pixel with obvious brightness changes in the digital images, eliminate irrelevant information, carry out feature extraction, and retain the important structural attributes of the images. Edge detection technology is also widely used in the preprocessing of astronomical spectrum images[11]. In our work, the Gaussian filter is used to smooth the spectrograms, reduce the background noise, and make the signal boundary clearer. Then different edge detection operators (Sobel, Scharr and Laplace Operator) are used to detect the edge of the smoothed spectrums to retain the important structure information of the signal contour.

## *2.4 Convolution Neural Network*

In recent years, great progress has been made in the research and development of deep learning, resulting in a series of supervised learning tasks, including image classification, image segmentation and target detection, which are close to or exceed human accuracy[12]. As the core of deep learning, convolution neural network (CNN), including VGG[13], ResNet[14], InceptionNet[15], MobileNet[16], DenseNet[17] and NASNet[18], has achieved excellent classification accuracy on popular image classification datasets. In this article, the authors use the pre-trained models provided by TensorFlow to load weights training on ImageNet[19], then modify the output dimensions of the fully connected layers, finally fine-tunes model parameters by transfer learning to achieve the classification task of SETI images. Transfer learning can effectively accelerate the training speed, obtain a better convergence final model.

## 3 Edge detection and deep learning based SETI signal classification

As for the noise reduction of spectrums, predecessors have tried many convolution methods at the pixel level, but they have not obtained satisfactory results. Low-pass, high-pass and band-pass filters are only applicable to the noise whose power spectral density is evenly distributed in the whole frequency domain, so they are not applicable to the spectrum of SETI, since the background of SETI spectrums are dominated by Gaussian noise with random noise intensity. In addition, the extremely low signal-to-noise ratio will make it impossible to clearly distinguish between the signal and background. Therefore, how to use appropriate preprocessing methods for SETI spectrum to reduce background noise and enhance the contour of signals has become the key problem to be solved in this paper.

Based on above analysis, this paper first converts spectrums in RGB into gray scale images, then convolutes and smoothes the images with Gaussian filter, using edge detection operator to enhance the contour of the signal shown on the images and make final experimental dataset. The transfer learning will be carried out on the pretrained models, and the Adamax optimizer[20] with adaptive learning rate is used to optimize the learning process and improve the performance of the classifier model. The framework of the model is shown below.

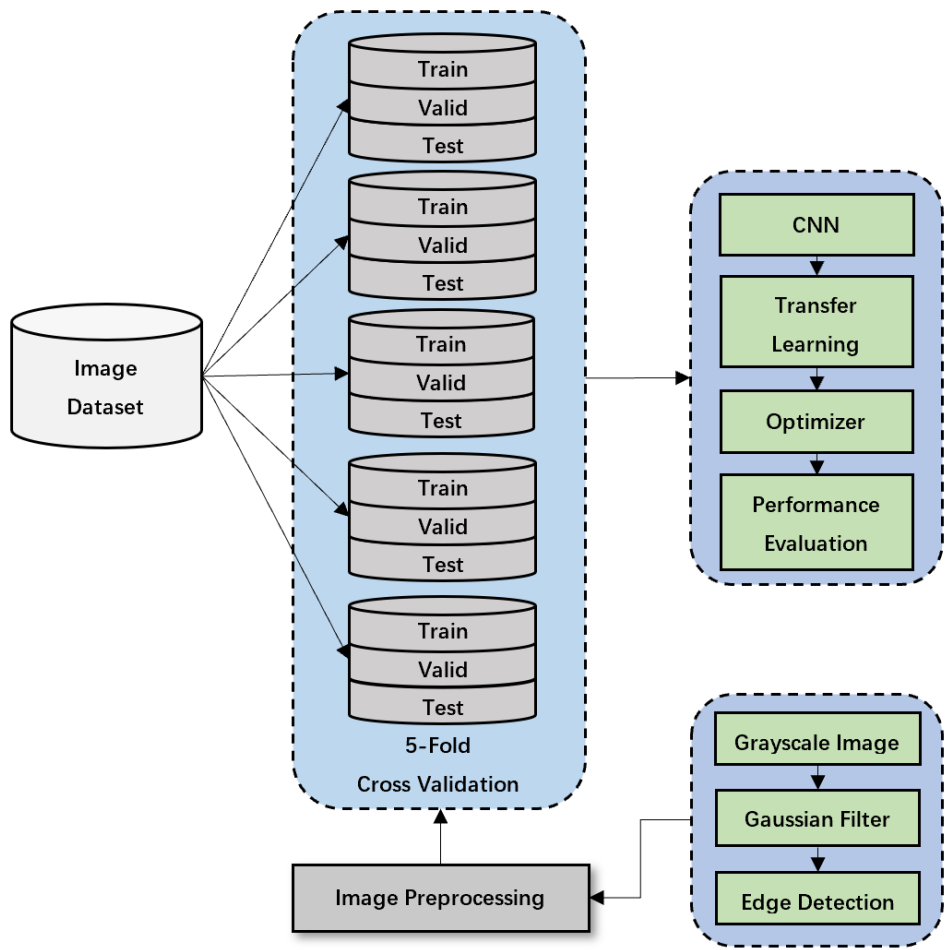

**Figure 2:** The framework of the model

### 3.1 Image Preprocessing

### 3.1.1 Gray Scale Image

After Fourier transform, signals generate colored spectrograms in RGB format. The size of an original image is 196608 pixels, and the resolution is 384x512. Because RGB is not necessary image information, OpenCV[21] is used to convert the image from BGR to gray scale image by image color conversion function after the image is read in BGR form. The grayscale image can compress the amount of data and facilitate subsequent processing and training. The pixel value of grayscale can also reflect the signal strength, which is convenient for obtaining the edge gradient information and the overall contour of the signals.

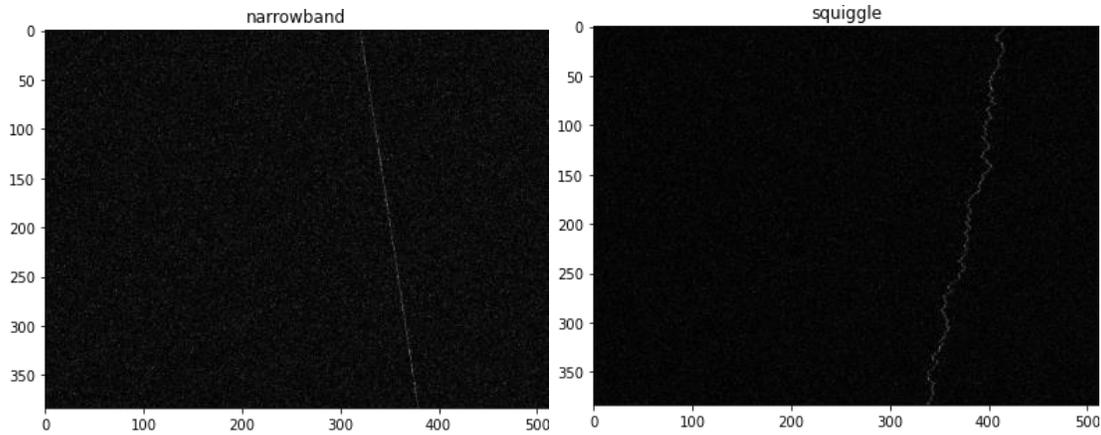

**Figure 3:** Narrowband and Squiggle signals are transformed into grayscale spectrums

### 3.1.2 Gaussian Filter

Gaussian filter is a kind of linear smoothing filter that is suitable for eliminating Gaussian noise and is widely used in the process of image denoising and smoothing. This paper uses a two-dimensional Gaussian convolution kernel with standard deviation of 1 and size of $3 \times 3$ to smoothes the spectrograms. Gaussian filter is similar to mean filter, but instead of using a simple mean. It uses a weighted mean, where neighborhood pixels that are closer to the central pixel contribute more weight to the average. The filter template can be calculated by two-dimensional Gaussian distribution, and the formula is as follows:

$$G(x, y) = \frac{1}{2\pi\sigma^2} e^{-\frac{x^2+y^2}{2\sigma^2}}$$
(3-1)

Gaussian filter is very effective in suppressing noise which obeys the normal distribution. Smoothing the spectrograms can reduce the background noise, make the signal boundary clearer, enhance the characteristics of image structure, and retain the overall grayscale distribution of the image, so as to obtain the image with high signal-to-noise ratio (SNR).

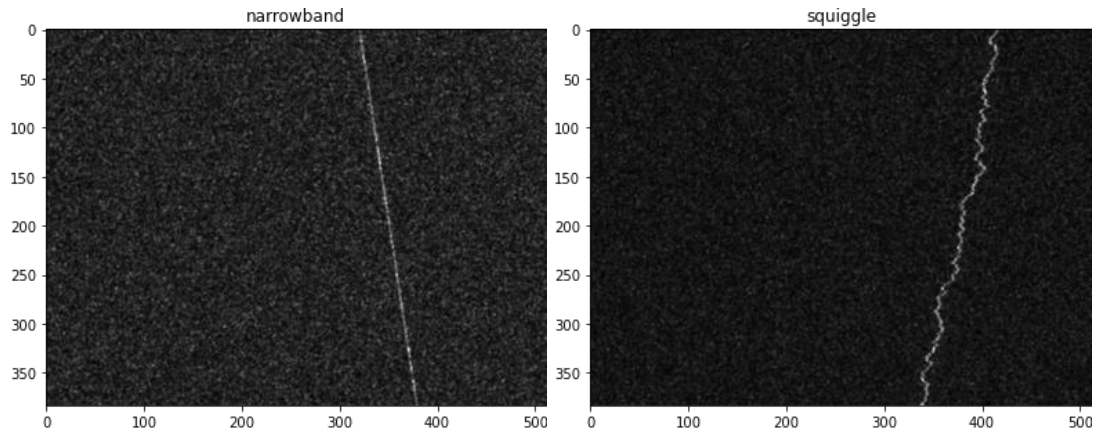

**Figure 4:** Gaussian smoothed Narrowband and Squiggle grayscale spectrums

*3.1.3 Edge Detection Operator*

Sobel operator detects the edge according to the phenomenon that the gray weighted difference between the upper, lower, left and right adjacent points of pixels reaches the extreme value at the edge. The operator can smooth the noise and provide more accurate edge direction information[22], but the positioning accuracy for edge is not high enough. Scharr operator is an improvement of Sobel operator[23]. In order to effectively extract weak edges, the difference between pixel values is increased by amplifying the weight coefficient in the filter. Both of Sobel and Scharr have the same principle and use mode in detecting edges. In addition, Laplace operator is a second-order differential linear operator. It detects the edge through the second-order derivative of the image. It has stronger edge positioning ability and better sharpening effect[24].

In order to further reduce the adverse interference of noise and retain the important structure information of the signal, this paper applied horizontal edge detection and vertical edge detection for the smoothed spectrum and compared the effects of different operators. After edge detection processing, the output images in horizontal and vertical directions are superimposed into a new image with a certain weight($\alpha$=0.5，$\beta$=0.5) to obtain a more accurate signal contour. In addition, the size of the edge detection filters used in this paper is 3×3, and the data type of the output images is 8-bit unsigned integer.

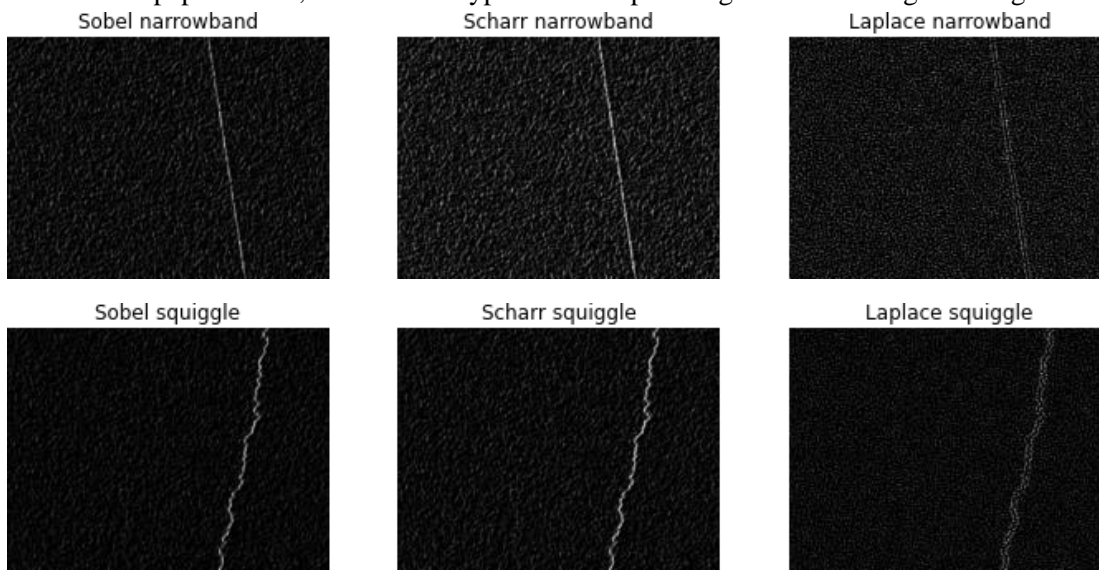

**Figure 5:** Preprocessed spectrums of Narrowband and Squiggle signals

### 3.2 Explore CNN Architecture

Convolutional neural network (CNN) is a kind of feedforward neural network with convolution calculation and deep structure. It is one of the most representative algorithms of deep learning, which greatly promotes the development of deep learning in the field of image recognition. We tested multiple convolutional neural network architectures in this classification task, using only spectral images as input data. The total size of the data set is 7000, including the spectrum images corresponding to seven different categories of SETI signals, and each category contains a total of 1000 pictures[4].

The CNN architecture tested in this study includes:

### 3.2.1 InceptionNet

Traditional networks increase the network depth through the superposition of layers, while InceptionNet improves the performance from the network width and proposes the Inception module. It puts multiple convolution or pooling operations together to assemble a network module. When designing the neural network, the whole network structure is assembled in the unit of these modules. InceptionNet maintains the computing cost while improving the adaptability to different scale images, increasing the depth and width of the architecture and the utilization of resources within the network[15].

### 3.2.2 Residual Network (ResNet)

In order to overcome the degradation of network performance caused by the gradient vanish and the network depth increasing during the training process, the residual network introduces the residual module into the CNN. The residual network is easy to optimize and can improve the accuracy by increasing a considerable depth. The internal residual module uses skip connection, which alleviates the gradient vanish problem caused by increasing depth in the deep neural network and allows the construction of a deeper network model[14].

### 3.2.3 Inception-ResNet

ResNet's network structure can speed up training and improve classification performance. Inception module can acquire sparse or non-sparse features on the same layer. Inception-ResNet, as the combination of ResNet and InceptionNet, achieves the complementary advantages of the two models, making the network model wide and deep with higher accuracy[25].

### 3.2.4 Densely Connected Residual Network (DenseNet)

DenseNet's basic idea is in line with ResNet's and proposes a more radical, dense connection mechanism: Connecting all layers together, specifically each layer accepts all the layers in front of it as its extra input[17]. DenseNet not only reduces the effect of gradient vanish on model training, but also enhances the transfer of features and reduces the number of parameters to a certain extent.

### 3.3 Transfer Learning

By means of transfer learning, this paper loads the pretrained model and gradually fine tunes to help the training of the new model. Since the parameters of excellent deep learning networks such as InceptionNet, ResNet, DenseNet have been well trained on the ImageNet dataset, so we directly use this parameter information as the initialization parameters of the new model we want to train. Then, according to the classification task in this paper, we reconstruct the fully connected layers based on the pretrained model, and then train the network.

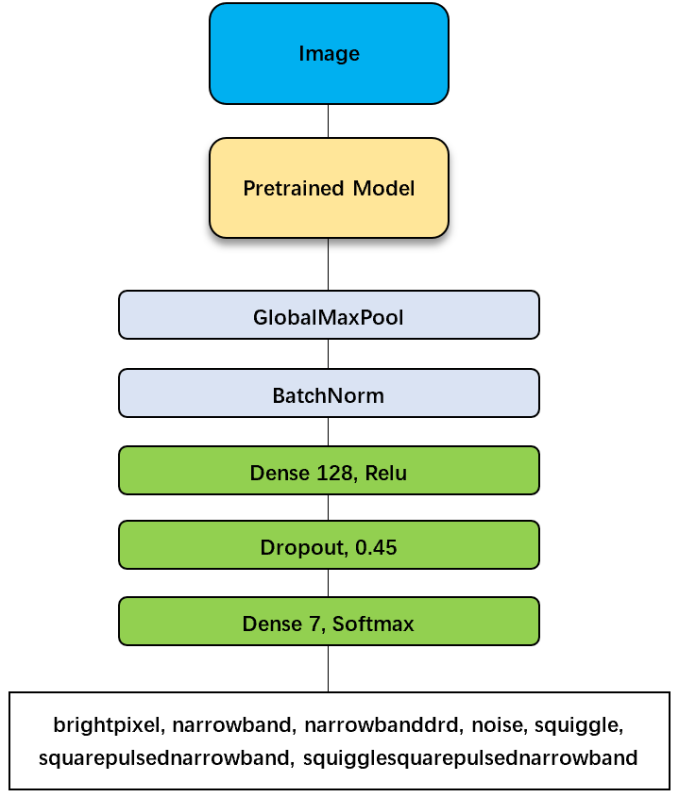

**Figure 6:** Image classification network based on Transfer Learning

### *3.4 Optimization Algorithm*

In the process of deep learning back propagation, in order to guide the parameters of the loss function to update the appropriate size in the right direction, so that the updated parameters make the value of the loss function approach the global minimum, we choose to use Adamax optimizer for optimization in this experiment. Adamax algorithm is a variant of Adam algorithm[20] based on infinite norm. It can make the algorithm of learning rate update more stable and simpler. The calculation formula of parameter update is as follows:

$$m_t = \mu * m_{t-1} + (1 - \mu) * g_t \tag{3-2}$$

$$n_t = \max(v * n_{t-1}, |g_t|) \tag{3-3}$$

$$\widehat{m_t} = \frac{m_t}{1 - \mu^t} \tag{3-4}$$

$$\widehat{n_t} = \frac{n_t}{1 - v^t} \tag{3-5}$$

$$\Delta\theta_t = -\frac{\widehat{m_t}}{\widehat{n_t} + \epsilon} * \eta \tag{3-6}$$

$m_t$ is the first moment estimation and $n_t$ is the second moment estimation, $\widehat{m_t}$ is used to correct the bias of $m_t$ and $\widehat{n_t}$ is used to correct the bias of $n_t$. The learning process of algorithm is dynamically adjusted according to gradient, $-\frac{\widehat{m_t}}{\widehat{n_t} + \epsilon}$ has dynamic constraints on the learning rate $\eta$ and it changes dynamically in a simple range. In a word, Adamax optimizer has little demand for memory and is suitable for high-dimensional space and large-scale samples. It can adaptively adjust the learning rate for different parameters. In the experiment, the value of $\mu$, $v$, $\eta$ are 0.9, 0.999 and 0.001 respectively, $\epsilon$ is used to ensure that the denominator is not zero and its value is $10e^{-8}$.

## 4. Experimental design and verification

### 4.1 Experimental Design

In order to verify whether the image preprocessing methods and various image classification networks mentioned above have good classification performance, we select the most representative and accurate image classification network algorithm in each category of InceptionNet, ResNet, InceptionResNet and DenseNet for comparative experiments with the combination of three different image preprocessing methods, including Sobel, Scharr, Laplace Operators (The original spectrograms were also tested as the control group).

After complete image preprocessing, the image dataset consists of 7 types of signals, with 1000 of each type, and the resolution of the input image is 128x256. To assess the robustness of the models, we used the five-fold cross validation method to divide 80% of the dataset into training sets and 20% into validation sets and test sets (10% for validation and test sets, respectively). Finally, the average number of Macro-F1 and Accuracy in the five cross-validations was calculated to evaluate the classification performance of the models.

### 4.2 Performance Evaluation Criteria

We used two evaluation criteria, F1 and Accuracy, which are commonly used in the field of machine learning to evaluate the performance of the trained model. These two evaluation criteria are defined by the confusion matrix (Table 1).

**Table 1:** Definition of Confusion Matrix

|        |          | Predicted           |                     |
|--------|----------|---------------------|---------------------|
|        |          | Positive            | Negative            |
| Actual | Positive | True Positive (TP)  | False Positive (FP) |
|        | Negative | False Negative (FN) | True Negative (TN)  |

#### 4.2.1 F1-score

F1-score is the weighted harmonic average of Precision and Recall. Precision indicates the actual number of positive samples in predicted positive samples, and Recall indicates the proportion of samples that are positive and also predict to be positive. The calculation formula of Precision, Recall and F1-score is shown in formula (4-1) (4-2) (4-3):

$$Precision = \frac{TP}{TP+FP} \quad\quad\quad (4\text{-}1)$$

$$Recall = \frac{TP}{TP+FN} \quad\quad\quad (4\text{-}2)$$

$$F1 = \frac{2\times Precision\times Recall}{Precision+Recall} \quad\quad\quad (4\text{-}3)$$

For the multi classification task in this paper, we extended the F1-score of the binary classification, using Macro-F1 as the measure in multi classification. We calculate Precision and Recall of each category to get their F1-score values, and then take the average value of them to get a Macro-F1 value.

#### 4.2.2 Accuracy

Accuracy refers to the ratio of the sample that correctly classified by the classifier to the total number of samples for a given test dataset, that is, the probability of correct prediction. The calculation is shown in formula (4-4):

$$Accuacy = \frac{TP+FN}{TP+FP+TN+FN} \quad\quad\quad (4\text{-}4)$$

### 4.3 Analysis of experimental results

**Table 2:** Comparison table of Macro-F1 and Accuracy of different models

| Model | | Accuracy | Macro-F1 | Parameter |
|---|---|---|---|---|
| InceptionV3 | Origin | 0.8140 | 0.8164 | 22,074,151 |
| | Sobel | **0.8960** | **0.8966** | |
| | Scharr | 0.8760 | 0.8778 | |
| | Laplace | 0.7912 | 0.7949 | |
| ResNet152V2 | Origin | 0.8083 | 0.8099 | 58,603,015 |
| | Sobel | **0.8932** | **0.8936** | |
| | Scharr | 0.8774 | 0.8786 | |
| | Laplace | 0.8163 | 0.8202 | |
| InceptionResNetV2 | Origin | 0.8072 | 0.8092 | 54,540,519 |
| | Sobel | **0.8929** | **0.8930** | |
| | Scharr | 0.8834 | 0.8843 | |
| | Laplace | 0.8037 | 0.8060 | |
| DenseNet201 | Origin | 0.8383 | 0.8412 | 18,576,455 |
| | Sobel | **0.9074** | **0.9082** | |
| | Scharr | 0.8932 | 0.8945 | |
| | Laplace | 0.8392 | 0.8371 | |

Table 2 demonstrates the performance comparison results of four different image classification networks combined with three image preprocessing methods. Macro-F1 and Accuracy are used in the table to measure the multi-classification accuracy and comprehensive performance of the model. It can be clearly seen that among four convolutional neural networks, DenseNet201 has better classification effect than InceptionV3, ResNet152V2 and InceptionResNetV2. Due to its compact architecture, DenseNet201 has fewer parameters, less memory and faster training.

Among three edge detection operators, Sobel has the best optimization effect, Scharr is the second, and Laplace operator is the worst and sometimes reduces the accuracy. This may be because Laplace operator is too sensitive to noise, which enhances the noise in the image and causes interference to the signals. In addition, the combined preprocessing algorithm of Grayscale, Gaussian filter and Sobel operator has optimized the image classification model well in terms of Macro-F1 and Accuracy. Compared with spectrums without preprocessing(Origin), the proposed preprocessing method highly improved the classification Accuracy.

The SETI signal classification model based on DenseNet201 combined with Sobel operator is obviously better than other combined algorithms, with 90.74% Accuracy and 0.9082 Macro-F1. Among all experimental results, the model with the best classification performance was also generated by DenseNet201 + Sobel, achieving the highest accuracy of 92.86 and 0.9310 Macro-F1. The performance scores of this model are shown as follows:

**Table 3:** Best model performance scores in the test set.

| | Precision | Recall | F1-score | Support |
|---|---|---|---|---|
| brightpixel | 0.9510 | 0.9151 | 0.9327 | 106 |
| narrowband | **0.9794** | **0.9794** | **0.9794** | 97 |
| narrowbanddrd | 0.9412 | 0.9524 | 0.9467 | 84 |

| noise | 0.7583 | 0.9579 | 0.8465 | 95 |
|---|---|---|---|---|
| squarepulsednarrowband | 0.9608 | 0.8376 | 0.8950 | 117 |
| squiggle | 0.9770 | 0.9659 | 0.9714 | 88 |
| squigglesquarepulsednarrowband | 0.9720 | 0.9204 | 0.9455 | 113 |
|  |  |  |  |  |
| accuracy |  |  | **0.9286** | 700 |
| macro avg | **0.9342** | **0.9327** | **0.9310** | 700 |

**Table 4:** Confusion matrix of best model performance scores.

| Predict \ Actual | bp | nb | nbdrd | no | sqpnb | sql | sqlspnb |
|---|---|---|---|---|---|---|---|
| bp | 97 | 0 | 0 | 8 | 1 | 0 | 0 |
| nb | 0 | 95 | 2 | 0 | 0 | 0 | 0 |
| nbdrd | 0 | 2 | 80 | 1 | 0 | 0 | 1 |
| no | 4 | 0 | 0 | 91 | 0 | 0 | 0 |
| sqpnb | 1 | 0 | 2 | 14 | 98 | 0 | 2 |
| sql | 0 | 0 | 0 | 3 | 0 | 85 | 0 |
| sqlspnb | 0 | 0 | 1 | 3 | 3 | 2 | 104 |

In Table 4, bp for brightpixel, nb for narrowband, nbdrd for narrowbanddrd, no for noise, sqpnb for squarepulsednarrowband, sql for squiggle, and sqlspnb for squigglesquarepulsednarrowband. And it can be analyzed from the table that the classification of noise has the lowest precision and F1-score, this is because if the signal-noise ratio (SNR) is too low, then signals will be covered by the background noise and misclassified as noise. And for narrowband signals, they have the highest precision, recall and F1-score. SETI project focuses on the the correct classification of narrowband signals[5], so it can be regarded as an advantage of this model. The classification of squarepulsednarrowband was also unsatisfied, which are often confused with noise. All in all, considering the relatively short training time and small memory usage, the proposed model can effectively improve the classification accuracy of SETI spectrums.

## 5 Conclusion

This paper proposed a new method to implement the classification of SETI simulated signals. Two-dimensional spectrum images are acquired using Fourier transform. Gaussian filter and edge detection operator are used to denoise and enhance the images. The classification accuracy of deep neural network is improved by the transfer learning and optimizer. The experiments conducted on datasets demonstrates that DenseNet201 combined with Gaussian filter and Sobel operator can improve the classification performance on SETI spectrograms and show its effectiveness. In addition, we will carry on further new research to denoise SETI spectrograms in a signal processing method which we haven't tried before, in order to achieve better classification accuracy.